\documentclass[english,a4paper]{article}
\usepackage{fullpage}
\usepackage{amsmath}
\usepackage{amssymb}
\usepackage{xcolor}
\usepackage{pdfcolmk}
\usepackage{url}
\usepackage{graphicx}
\usepackage{listings}
\PassOptionsToPackage{normalem}{ulem}
\usepackage{ulem}
\usepackage[unicode=true,
 bookmarks=false,
 breaklinks=false,pdfborder={0 0 1},backref=section,colorlinks=false]
 {hyperref}
 \usepackage[square,numbers]{natbib}

\makeatletter

\providecolor{lyxadded}{rgb}{0,0,1}
\providecolor{lyxdeleted}{rgb}{1,0,0}

\DeclareRobustCommand{\lyxsout}[1]{\ifx\\#1\else\sout{#1}\fi}

\usepackage{lineno}
\modulolinenumbers[5]
\usepackage{dsfont}
\usepackage{float}
\floatstyle{ruled}
\newfloat{code}{H}{loa}
\floatname{code}{Code}

\makeatother

\title{Automated learning with a probabilistic programming language: Birch}

\author{Lawrence M. Murray \\
Uppsala University \and Thomas B. Sch\"on \\
Uppsala University}
\date{}

\begin{document}
\maketitle

\begin{abstract}
This work offers a broad perspective on probabilistic modeling and
inference in light of recent advances in \emph{probabilistic programming},
in which models are formally expressed in Turing-complete programming
languages. We consider a typical workflow and how probabilistic programming
languages can help to automate this workflow, especially in the matching
of models with inference methods. We focus on two properties of a
model that are critical in this matching: its \emph{structure}---the
conditional dependencies between random variables---and its \emph{form}---the
precise mathematical definition of those dependencies. While the structure
and form of a probabilistic model are often fixed \textsl{a priori},
it is a curiosity of probabilistic programming that they need not
be, and may instead vary according to random choices made during program
execution. We introduce a formal description of models expressed as
programs, and discuss some of the ways in which probabilistic programming
languages can reveal the structure and form of these, in order to
tailor inference methods. We demonstrate the ideas with a new probabilistic
programming language called \emph{Birch}, with a multiple object tracking
example.
\end{abstract}

\section{Introduction}

Probabilistic approaches have become standard in system identification,
machine learning and statistics, particularly in situations where
the quantification of uncertainty or assessment of risk is paramount.
A typical workflow proceeds through several stages, from experimental
design, to data collection, to model development, to prior elicitation,
to inference, to decision making. At least part of this workflow involves
computer code. For the inference stage, this is often bespoke code
tailored to a particular study: it couples the implementation of the
model with the implementation of the chosen inference method. The
model code is not easy to reuse with a different method, nor the method
code with a different model.

As data size and compute capacity increase, the complexity of models,
and their implementations, increases too. Complex models arise in
numerous fields, including nonparametrics, where there is an unbounded
number of variables, in object tracking and phylogenetics, where data
structures such as random finite sets and random trees appear, and
in numerical weather prediction and oceanography, where specialized
numerical methods are used for continuous-time systems and partial
differential equations. For such complex models, bespoke implementations
may involve nontrivial---even tedious---manual work, such as deriving
the full conditionals of the posterior distribution, or calculating
the gradients of a complex likelihood function, or tuning numerical
methods for stability. The effort may need repeating if the model
or inference method is later changed.

A more scalable approach to implementation is desirable. Recognizing
this, there has been a tradition of software that separates model
specification from method implementation for the purposes of inference
(e.g. WinBUGS~\cite{Lunn2000}, OpenBUGS~\cite{Lunn2012}, JAGS~\cite{JAGS,Plummer2003},
Stan~\cite{STAN}, Infer.NET~\cite{InferNET18}, LibBi~\cite{Murray2015},
Biips~\cite{Todeschini2014}). Typically, this software supports
one predominant method but many possible models. The methods include
Markov chain Monte Carlo (MCMC) methods such as the Gibbs sampler
for WinBUGS, OpenBUGS and JAGS, and Hamiltonian Monte Carlo (HMC)
for Stan, variational methods for Infer.NET, and Sequential Monte
Carlo (SMC) methods for LibBi and Biips. The software provides a way
to adapt the method for a large number of models and automate routine
procedures, such as adaptation of Markov kernels in WinBUGS~\cite[p. 6]{Spiegelhalter2003},
and automatic differentiation in Stan. It is typical for these languages
to restrict the set of probabilistic models that can be expressed,
in order to provide an inference method that works well for this restricted
set. Stan, for example, works only with differentiable models using
HMC, while LibBi works only with state-space models using SMC-based
methods. Models outside of these sets may require more specialist
tools. In phylogenetics, for example, RevBayes~\cite{Hoehna2016}
provides the particular modeling feature of \emph{tree plates} to
represent phylogenetic trees, for which specialized Markov kernels
can be applied within MCMC.

Naturally, methods have also become more complex to accommodate these
more complex models and larger data sets. Modern Monte Carlo methods
often nest multiple baseline algorithms, such as SMC within MCMC,
as in particle MCMC~\cite{Andrieu2010}, or SMC within SMC, as in
SMC$^{2}$~\cite{Chopin2013}. Data subsampling-based algorithms~\cite{Bardenet2017}
are becoming standard for dealing with large data sets. Monte Carlo
samplers increasingly use gradient information, such as the Metropolis-adjusted
Langevin algorithm (MALA)~\cite{Roberts1998}, HMC~\cite{Neal2011a},
and deterministic piecewise samplers~\cite{Bouchard-Cote2017,Bierkens2016,Vanetti2017}.
Various methods manipulate the stream of random numbers (e.g. \cite{Murray2013a,Gerber2015,Deligiannidis2016})
or potential functions (e.g. \cite{Whiteley2014,DelMoral2015}) to
improve estimates. Software has begun to address this complexity in
methods, too. NIMBLE~\cite{deValpine2017}, for example, uses models
similar to those of WinBUGS, OpenBUGS and JAGS, but provides manual
customization of the Markov kernels used within MCMC.

We see value in flexible tools that allow for the implementation of
both complex models and complex methods, and in moving from one-to-many
tools (one method, many models) to many-to-many tools (many methods,
many models). To this end, we consider the potential of \emph{probabilistic
programming}: a programming paradigm that aims to accelerate workflow
with new programming languages and software tools tailored for probabilistic
modeling and inference. In particular, it aims to develop Turing-complete
programming languages for model implementation, extending existing
languages for model specification with programming concepts such as
conditionals, recursion (loops), and higher-order functions, for greater
expressivity. It aims to \emph{decouple} the implementation of models
and methods into modular components that can be reassembled and reused
in multiple configurations in a many-to-many manner. It aims to automate
the selection of an inference method for a given model, and to automate
the tuning necessary for it to work well in practice. These goals
remain aspirational, and an active area of research across disciplines
including machine learning, statistics, system identification, artificial
intelligence and programming languages.

A number of probabilistic programming languages have been developed
with such aims in recent years. Examples include Church~\cite{Goodman2008},
BLOG~\cite{Milch2007}, Venture~\cite{Mansinghka2014}, WebPPL~\cite{Goodman2014},
Anglican~\cite{Tolpin2016}, Figaro~\cite{Pfeffer2016}, Turing~\cite{Ge2018},
Edward~\cite{Tran2016}, and Pyro~\cite{Pyro}. These all explore
different approaches that reflect, in the first instance, the different
problem domains to which they are orientated, and in the second instance,
the relatively young age of the field. All of these languages are
considered \emph{universal} probabilistic programming languages---i.e.
Turing-complete programming languages that admit arbitrary models
rather than restricted sets. This is not to say, of course, that efficient
inference is possible for all models that are admitted---but these languages
can work well for a large class of models for which they do have efficient
inference methods, they do provide useful libraries for implementing
probabilistic models, and they may support the development or customization
of inference methods from within the same language.

We introduce a new universal probabilistic programming language called
\emph{Birch} (\url{www.birch-lang.org}), which implements the ideas
presented in this work. Birch is an imperative language geared toward
object-oriented and generic programming paradigms. It draws inspiration
from several sources, notably from LibBi~\cite{Murray2015}---for which
it is something of a successor---but in moving from model specification
language to universal probabilistic programming language it draws
ideas from modern object-oriented programming languages such as Swift,
too. Birch is Turing complete, with control flow statements such as
conditionals and loops, support for unbounded recursion and higher-order
functions, and dynamic memory management. Birch code compiles to C++14
code, providing ready access to the established ecosystem of C/C++
libraries available for scientific and numeric computing. A key component
of Birch is its implementation of delayed sampling~\cite{Murray2018},
a heuristic to provide optimizations via partial analytical solutions
to inference problems. While broadly applicable across problem domains,
invariably the approach taken in Birch is flavored by the perspective
of its developers, and so by applications in statistics, machine learning
and system identification.

This work is intended as a ``big picture'' perspective on the probabilistic
workflow, and how new ideas in probabilistic programming can assist
this. Birch is the concrete manifestation of these ideas. Throughout,
we make use of the state-space model as a running example. While the
ideas presented are not restricted to such models, they concretely
illustrate some of the core concepts, and have numerous practical
applications. In Section \ref{sec:models} we introduce a formal description
of the class of models considered in probabilistic programming. In
Section \ref{sec:methods} we introduce some methods of inference
and consider some of the ways in which probabilistic programming languages
can automate the many-to-many matching of models with inference methods.
In Section \ref{sec:demonstration} we introduce the Birch probabilistic
programming language as a specific implementation of these ideas.
In Section \ref{sec:worked-example} we work through the concrete
example of a multiple object tracking model---a state-space model with
random size---and show how it is implemented in Birch, and the inference
results obtained. Finally, we summarize in Section \ref{sec:discussion}.

\section{Models expressed in a programming language\label{sec:models}}

Probabilistic programming considers models expressed in Turing-complete
programming languages. Such models are usually referred to as \emph{probabilistic
programs}---qualified to \emph{universal probabilistic programs} when
one wishes to regard only the broadest class in terms of expressivity---and
described in programming language nomenclature. Here, we adopt probabilistic
nomenclature instead, to provide a more accessible treatment for the
intended audience. Taking the lead from the term \emph{graphical model}---a
model expressed in a graphical language---we suggest that the term \emph{programmatic
model}---a model expressed in a programming language---might be more appropriate
for this audience, and adopt this term throughout. Specifically, we
avoid the use of the term \emph{program} when referring only to a
model implementation, as in ordinary usage one thinks of a computer
program as combining the implementation of both a model and an inference
method, which can cause confusion. The term can also be misleading
given unrelated but similarly-named concepts in system identification,
such as linear programs and stochastic programs.

We follow the statistics convention of using uppercase letters to
denote random variables (e.g. $V$) and lowercase letters to denote
instantiations of them (e.g. $v$), with $v\in\mathbb{V}$. We then
adopt measure theory notation to clearly distinguish between distributions
(which we will ultimately simulate) and likelihood functions (which
we will ultimately evaluate): the distribution of a random variable
$V$ is denoted $p(\mathrm{d}v)$, while evaluation of an associated
probability density function (pdf, for continuous-valued random variables)
or probability mass function (pmf, for discrete-valued random variables)
is denoted $p(v)$.

Assume that we have a countably infinite set of random variables $\{V_{k}\}_{k=1}^{\infty}$,
with a joint probability distribution over them, which has been implemented
in code in some programming language. The only stochasticity available
to the code is via these random variables. We execute the code, and
as it runs it encounters a finite subset of the random variables in
some order determined by that code. Denote this order by a permutation
$\sigma$, with its (random) length denoted $|\sigma|$, defining
a sequence $(V_{\sigma[k]})_{k=1}^{|\sigma|}$. The first element,
$\sigma[1]$, is always the same. Each subsequent element, $\sigma[k]$,
is given by a function of the random variables encountered so far,
denoted $\mathrm{Ne}$ (for \emph{next}), so that $\sigma[k]=\mathrm{Ne}(v_{\sigma[1]},\ldots,v_{\sigma[k-1]})$.
This function $\mathrm{Ne}$ is implied by the code. Note that $\mathrm{Ne}$
is a deterministic function given preceding random variates, as there
is no stochasticity available to the code except via these. This is
also why $\sigma[1]$ is always the same: no source of stochasticity
precedes it.

As each random variable $V_{\sigma[k]}$ is encountered, the code
associates it with a distribution
\[
V_{\sigma[k]}\sim p_{\sigma[k]}\left(\mathrm{d}v_{\sigma[k]}\mid\mathrm{Pa}(v_{\sigma[1]},\ldots,v_{\sigma[k-1]})\right),
\]
where $\mathrm{Pa}$ (for \emph{parents}) is a deterministic function
of the preceding random variates, selecting from them a subset on
which the distribution of $V_{\sigma[k]}$ depends. It is possible
that the distribution $p_{\sigma[k]}$ also depends on exogenous factors
such as user input; we leave this implicit to simplify notation.

At some point the execution terminates, having established the distribution
\[
p_{\sigma}(\mathrm{d}v_{\sigma[1]},\ldots,\mathrm{d}v_{\sigma[|\sigma|]})=\prod_{k=1}^{|\sigma|}p_{\sigma[k]}\left(\mathrm{d}v_{\sigma[k]}\mid\mathrm{Pa}(v_{\sigma[1]},\ldots,v_{\sigma[k-1]})\right).
\]

We will execute the code several times. The $n$th execution will
be associated with the distribution $p_{\sigma_{n}}$, given by
\[
p_{\sigma_{n}}(\mathrm{d}v_{\sigma_{n}[1]},\ldots,\mathrm{d}v_{\sigma_{n}[|\sigma_{n}|]})=\prod_{k=1}^{|\sigma_{n}|}p_{\sigma_{n}[k]}\left(\mathrm{d}v_{\sigma_{n}[k]}\mid\mathrm{Pa}(v_{\sigma_{n}[1]},\ldots,v_{\sigma_{n}[k-1]})\right),
\]
with $\sigma_{n}[k]=\mathrm{Ne}(v_{\sigma_{n}[1]},\ldots,v_{\sigma_{n}[k-1]})$.
Subscript $n$ is used to denote execution-dependent variables. For
different executions $n$ and $m$, it is possible for the number
of random variables encountered ($|\sigma_{n}|$ and $|\sigma_{m}|$)
to differ, for the sequences of random variables $(V_{\sigma_{n}[k]})_{k=1}^{|\sigma_{n}|}$
and $(V_{\sigma_{m}[k]})_{k=1}^{|\sigma_{m}|}$ to differ, and even
for the two subsets of random variables $\{V_{\sigma_{n}[k]}\}_{k=2}^{|\sigma_{n}|}$
and $\{V_{\sigma_{m}[k]}\}_{k=2}^{|\sigma_{m}|}$ to be disjoint (recall
that the first random variable to be encountered is always the same).
In general, we should therefore assume that $p_{\sigma_{n}}$ and
$p_{\sigma_{m}}$ are not the same, but rather components of a larger
mixture.

The above describes the class of models that we refer to as \emph{programmatic
models}. The permutation $\sigma_{n}$ reflects the fact that a program
can make conditional choices during execution that are based on the
simulation of random variables, and that these may lead to very different
outcomes. Consider, for example, a model implementation that begins
with a coin flip: on heads it executes one model, on tails some other
model. Such an implementation represents a mixture of two models,
but each execution can encounter only one.

We are interested in two properties of a programmatic model from an
inference perspective:
\begin{enumerate}
\item \emph{structure}, by which we mean the factorization of the joint
distribution $p_{\sigma_{n}}$ into conditional distributions $p_{\sigma_{n}[1]},\ldots,p_{\sigma_{n}[|\sigma_{n}|]}$,
and
\item \emph{form}, by which we mean the precise mathematical definition
of the conditional distributions $p_{\sigma_{n}[1]},\ldots,p_{\sigma_{n}[|\sigma_{n}|]}$.
\end{enumerate}

\subsection{Structure}

The structure of a probabilistic model is typically defined as the
dependency relationships between random variables. Popular model classes
such as hidden Markov models (HMMs), state-space models (SSMs), Markov
random fields, etc, encode particular structures for specialist purposes
such as dynamical systems and spatial systems. Generalizing these,
structure is perhaps most explicitly encoded by \emph{graphical models}
(see e.g. \cite{Jordan2004,Bishop2007,Koller2009}), where a probabilistic
model is represented as a graph, with nodes as random variables, and
edges encoding the relationships between them. Generic inference techniques
such as the sum-product algorithm~\cite{Pearl1988} make explicit
use of the graph---and thus the structure of the model---to perform inference
efficiently with respect to the number of computations required.

We are interested in the same for programmatic models. For illustration,
we can readily compare the class of programmatic models to the class
of \emph{directed graphical models}. Like all graphical models, the
nodes of a directed graphical model represent random variables. The
edges are directed and represent a conditional dependency relationship
from a \emph{parent} at the tail of the arrow, to a \emph{child} at
the head of the arrow. The entire graph must be acyclic.

We can represent directed graphical models as programmatic models
within the formal definition above. The nodes and edges are known
\textsl{a priori} and establish a joint distribution, over $K$ random
variables, of:
\[
p(\mathrm{d}v_{1},\ldots,\mathrm{d}v_{K})=\prod_{k=1}^{K}p_{k}(\mathrm{d}v_{k}\mid\mathrm{pa}_{k}),
\]
where we use $\mathrm{pa}_{k}$ to denote the set of parents of $V_{k}$
under the graph. The model implementation in code must necessarily
encounter the nodes in a valid topological ordering given the edges.
On each execution $n$, the same finite set of random variables $\{V_{k}\}_{k=1}^{K}$
is encountered. A directed edge from $V_{i}$ to $V_{j}$ indicates
that, if $V_{j}$ is the $k$th random variable to be encountered,
then $V_{i}$ must necessarily have been encountered already, and
$v_{i}\in\mathrm{Pa}(v_{\sigma_{n}[1]},\ldots,v_{\sigma_{n}[k-1]})$.
The conditional distribution assigned to any $V_{k}$ is always the
same $p_{k}(\mathrm{d}v_{k}\mid\mathrm{pa}_{k})$. Consequently, the
execution establishes the distribution:
\begin{align}
p_{\sigma_{n}}(\mathrm{d}v_{\sigma_{n}[1]},\ldots,\mathrm{d}v_{\sigma_{n}[|\sigma_{n}|]}) & =\prod_{k=1}^{|\sigma_{n}|}p_{\sigma_{n}[k]}\left(\mathrm{d}v_{\sigma_{n}[k]}\mid\mathrm{Pa}(v_{\sigma_{n}[1]},\ldots,v_{\sigma_{n}[k-1]})\right)\nonumber \\
 & =\prod_{k=1}^{K}p_{k}(\mathrm{d}v_{k}\mid\mathrm{pa}_{k})\nonumber \\
 & =p(\mathrm{d}v_{1},\ldots,\mathrm{d}v_{K}).\label{eq:directed-graphical-model}
\end{align}
This is to say that, while executions may encounter the random variables
in different orders, according to how the directed graphical model
has been implemented, each execution will always encounter the same
finite subset of $K$ variables, and establish the same structure
and form. If this is not the case, then the code must not be a correct
implementation of the directed graphical model that was given.

This motivates the difficulty of dealing with a programmatic model:
unlike for a directed graphical model, the structure of a programmatic
model is not known \textsl{a priori}. If a programmatic model were
expressed as a graph, the nodes of the graph would not be known until
revealed through the function $\mathrm{Ne}$, and only at that same
time would incoming edges to the node be revealed through the function
$\mathrm{Pa}$. The model must be executed to discover its structure.
But, furthermore, the nodes and edges may differ between executions,
so it is not simply a matter of executing the program once to determine
the complete structure.

\begin{figure}[t]
\begin{minipage}[t][1\totalheight][c]{0.32\columnwidth}%
\includegraphics[width=0.9\textwidth]{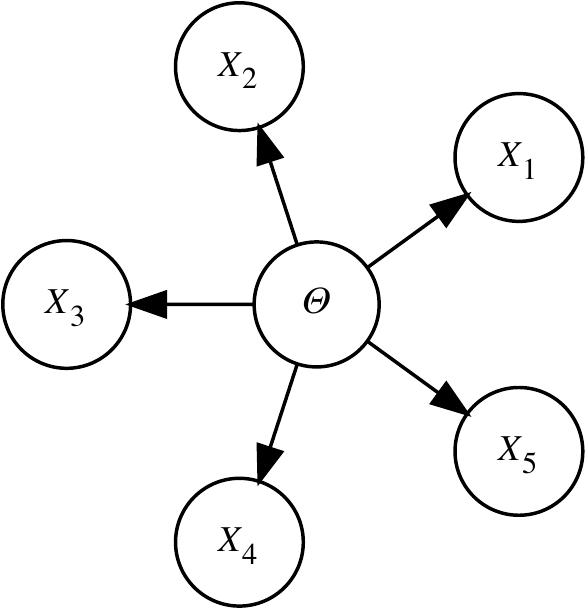}%
\end{minipage}%
\begin{minipage}[t][1\totalheight][c]{0.42\columnwidth}%
\includegraphics[width=0.9\textwidth]{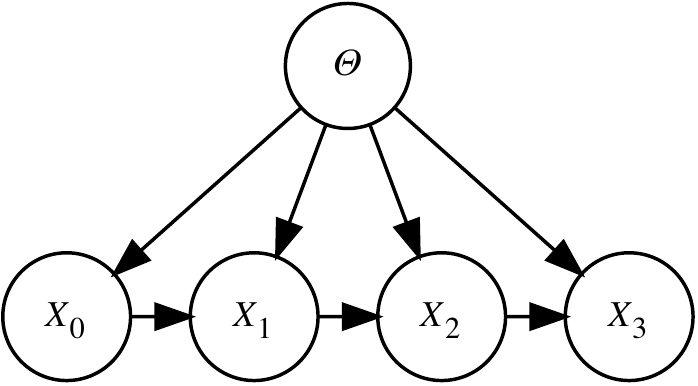}%
\end{minipage}%
\begin{minipage}[t][1\totalheight][c]{0.25\columnwidth}%
\includegraphics[width=0.9\textwidth]{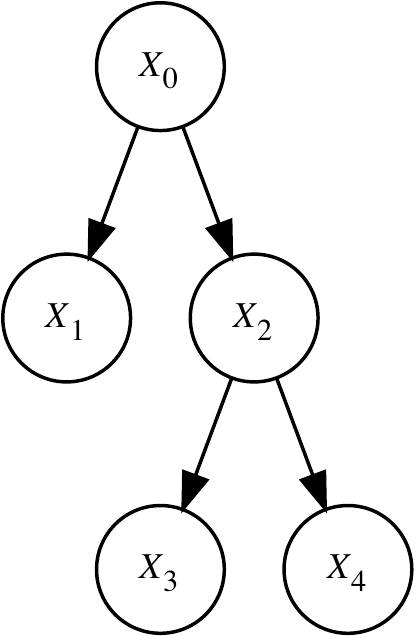}%
\end{minipage}

\vspace{3mm}

\begin{minipage}[t]{0.5\columnwidth}%
\includegraphics[width=0.9\textwidth]{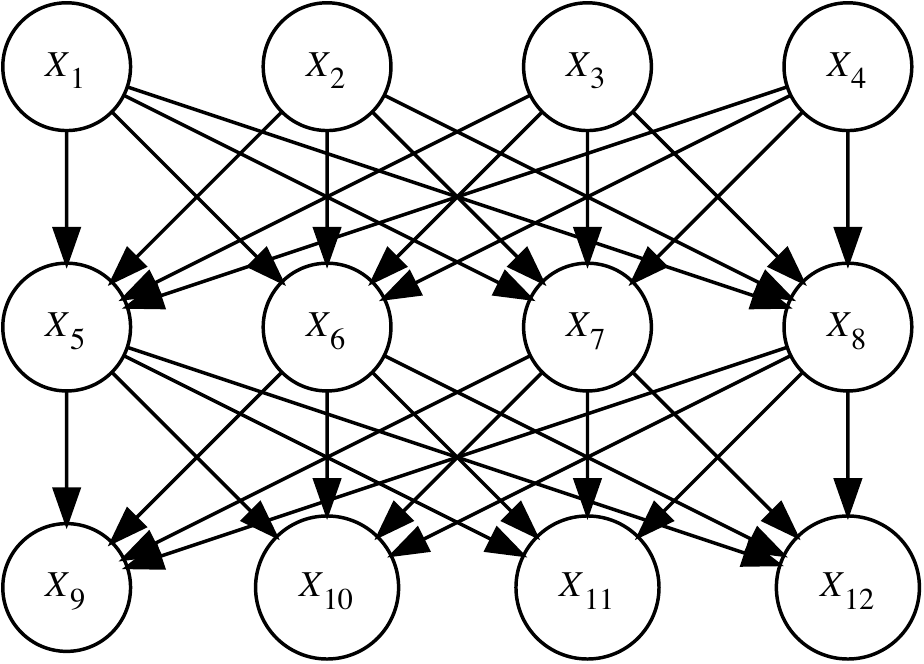}%
\end{minipage}%
\begin{minipage}[t]{0.5\columnwidth}%
\includegraphics[width=0.9\textwidth]{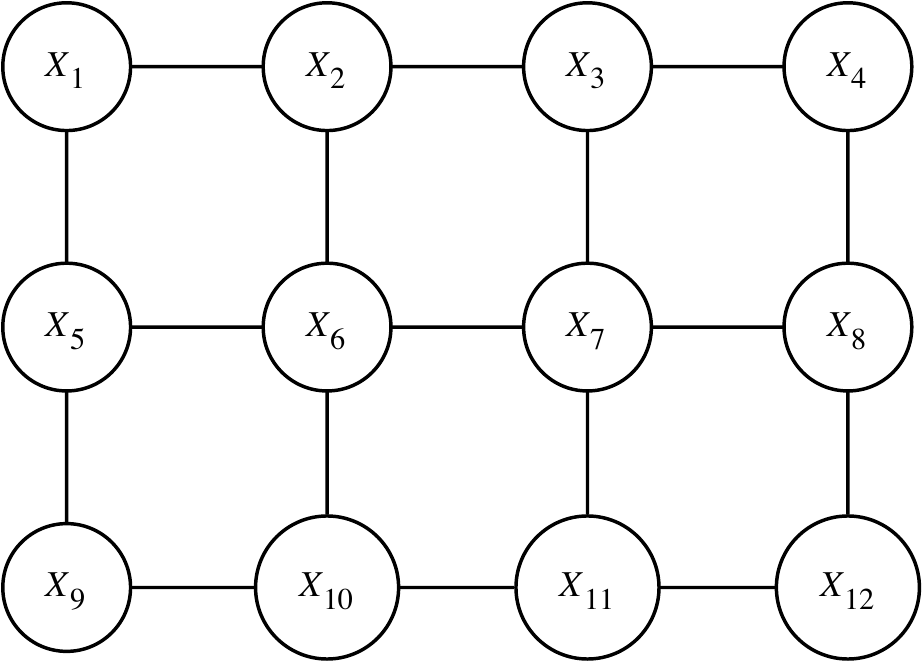}%
\end{minipage}

\caption{Structural motifs that occur frequently in probabilistic models, each
with a global parameter. Clockwise from top-left: star, chain, tree,
layer, grid.\label{fig:motifs}}
\end{figure}

We should not get too caught up on the general at the expense of the
useful, however. Directed graphical models are useful, and many models
used in practice can be expressed this way---it is those that cannot
be expressed this way that motivate the more general treatment of
programmatic models. Probabilistic models tend not be entangled assemblies
of random variables connected in arbitrary ways, but rather arranged
in recursive substructures such as chains, grids, stars, trees, and
layers. We refer to these as structural \emph{motifs} (see Figure
\ref{fig:motifs}). These motifs occur so frequently in probabilistic
models that there have been attempts to automatically learn model
structure based on them~\cite{Ellis2013}. For example, the chain
motif is the dominant feature of HMMs and SSMs, the grid motif that
of spatial models, the tree motif that of phylogenetic trees or stream
networks, the layer motif that of neural networks. The star motif
occurs whenever there are repeated observations of some system, for
example repeated time series observations of the same dynamical system,
where the parameters are identified across time series, or multiple
individuals in a medical study sharing common parameters. Parameters
with global influence are also a common occurrence, determining the
conditional probability distributions over variables in a chain or
grid, for example.

These motifs can be used to characterize a model, especially for the
purposes of selecting inference methods. While the nature of a programmatic
model is that its structure may change between executions, it may
be the case that a particular motif persists between executions, and
that this is known \textsl{a priori}. If we can characterize this
motif, we can leverage specialized inference methods for it, while
reserving generalized inference methods for the remainder of the structure.
We return to this idea in Section~\ref{sec:demonstration}.

\subsection{Form}

The form of a probabilistic model refers to the mathematical definition
of its distributions. In the case of programmatic models, this refers
to the conditional probability distributions $p_{\sigma_{n}[1]},\ldots,p_{\sigma_{n}[|\sigma_{n}|]}$.

In many cases these are common parametric forms such as Gaussian,
Poisson, binomial, beta and gamma distributions, with parameters given
by the parents of the random variable. Parametric distributions such
as these are readily simulated using standard algorithms (see e.g.
\cite{Devroye1986}), and admit either a pdf or pmf that can be evaluated,
and perhaps differentiated with respect to its parameters. More difficult
forms are those that can be simulated or evaluated but not both. For
example, a nonlinear diffusion process defines a distribution $p(\mathrm{d}x(t+\Delta t)\mid x(t))$
that can be simulated (at least numerically), but it may be prohibitively
expensive to evaluate the associated pdf $p(x(t+\Delta t)\mid x(t))$
for given values $x(t+\Delta t)$ and $x(t)$, or this may have no
closed form. Conversely, the classic Ising model is defined as a product
of potentials that readily permits evaluation of the likelihood of
any state, but requires expensive iterative computations to simulate.

Form may also carry information across structure. For example, where
the form of the parents of a random variable is conjugate to the form
of that random variable, a conjugate prior relationship is established.
In such cases, analytical marginalization and conditioning optimizations
may be possible within an inference method.

\subsection{Example}

We demonstrate how these abstract ideas apply to the concrete case
of an SSM. The SSM consists of a latent process $(X_{t})_{t=1}^{T}$,
observed process $(Y_{t})_{t=1}^{T}$, and optional parameters $\Theta$.
The joint probability distribution is given by:
\begin{equation}
p\left(\mathrm{d}y_{1:T},\mathrm{d}x_{1:T},\mathrm{d}\theta\right)=\underbrace{p\left(\mathrm{d}\theta\right)}_{\text{parameter}}\underbrace{p\left(\mathrm{d}x_{1}\mid\theta\right)}_{\text{initial}}\prod_{t=2}^{T}\underbrace{p\left(\mathrm{d}x_{t}\mid x_{t-1},\theta\right)}_{\textrm{transition}}\prod_{t=1}^{T}\underbrace{p\left(\mathrm{d}y_{t}\mid x_{t},\theta\right)}_{\text{observation}},\label{eq:SSMjoint}
\end{equation}
where we have assigned common names to the various conditional probability
distributions that make up this joint.

There are alternative representations. In the engineering literature,
it is common to represent dependencies between the random variables
using deterministic functions plus noise: \label{eq:SSMengineering}
\begin{align*}
x_{t} & =f(x_{t-1},\theta)+\xi_{t}, & y_{t} & =g(x_{t},\theta)+\zeta_{t}.
\end{align*}
Here, the parameter and initial distributions are as per \eqref{eq:SSMjoint},
but now the transition and observation distributions are expressed
via the (possibly nonlinear) functions $f$ and $g$, plus independent---often
Gaussian---noise terms $\xi_{t}$ and $\zeta_{t}$: \begin{subequations}
\begin{align}
\underbrace{p(\mathrm{d}x_{t}\mid x_{t-1},\theta)}_{\text{transition}} & =p_{\xi_{t}}(\mathrm{d}x_{t}-f(x_{t-1},\theta))\\
\underbrace{p(\mathrm{d}y_{t}\mid x_{t},\theta)}_{\text{observation}} & =p_{\zeta_{t}}(\mathrm{d}y_{t}-g(x_{t},\theta)),
\end{align}
\end{subequations} where $p_{\xi_{t}}$ denotes the distribution
of the process noise $\xi_{t}$, and $p_{\zeta_{t}}$ the distribution
of the observation noise $\zeta_{t}$.

This is a mathematical description of the standard SSM. To understand
it as a programmatic model, denote the set of random variables as
$\{V_{k}\}_{k=1}^{2T+1}=\{\Theta,X_{1},\ldots,X_{T},Y_{1},\ldots,Y_{T}\}$.
In this case the set is finite (or, equivalently, the infinite complement
of the set is never encountered). An implementation of the model in
Birch may look like the following (variable and function declarations
have been removed for brevity):\begin{code}
\begin{lstlisting}[mathescape]
$\theta$ $\sim$ Uniform(0.0, 1.0);
x[1] $\sim$ Gaussian(0.0, 1.0);
y[1] $\sim$ Gaussian(g(x[1], $\theta$), 0.1);
for t in 2..T {
  x[t] $\sim$ Gaussian(f(x[t-1], $\theta$), 1.0);
  y[t] $\sim$ Gaussian(g(x[t], $\theta$), 0.1);
}
\end{lstlisting}
\noindent \caption{\label{code:model}}
 \end{code}

\noindent Recall that $\mathrm{Ne}$ is the function that denotes
the next random variable to be encountered given the values of those
encountered so far, and $\mathrm{Pa}$ is the function that denotes
the parents of that random variable, given the same values. If we
think through executing the above code line-by-line we see that, for
example, $\mathrm{Ne}(\theta,x_{1:t-1},y_{1:t-1})=X_{t}$ and $\mathrm{Pa}(\theta,x_{1:t-1},y_{1:t-1})=\{\theta,x_{t-1}\}$;
also $\mathrm{Ne}(\theta,x_{1:t},y_{1:t-1})=Y_{t}$ and $\mathrm{Pa}(\theta,x_{1:t},y_{1:t-1})=\{\theta,x_{t}\}$.
For some execution $n$, the order of random variables encountered,
$\sigma_{n}$, will be:
\begin{align*}
\sigma_{n}[1] & =1 & \text{i.e. } & \Theta\\
\sigma_{n}[2] & =2 & \text{i.e. } & X_{1}\\
\sigma_{n}[3] & =T+2 & \text{i.e. } & Y_{1}\\
\sigma_{n}[5] & =3 & \text{i.e. } & X_{2}\\
\sigma_{n}[6] & =T+3 & \text{i.e. } & Y_{2}\\
\ldots\text{etc}
\end{align*}
Now consider an alternative implementation: \begin{code}
\begin{lstlisting}[mathescape]
$\theta$ $\sim$ Uniform(0.0, 1.0);
x[1] $\sim$ Gaussian(0.0, 1.0);
for t in 2..T {
  x[t] $\sim$ Gaussian(f(x[t-1], $\theta$), 1.0);
}
for t in 1..T {
  y[t] $\sim$ Gaussian(g(x[t], $\theta$), 0.1);
}
\end{lstlisting}
\caption{\label{code:alternative-implementation}}
\end{code}

\noindent This expresses precisely the same mathematical model, but
in a different programmatic form. When the code is executed, the order
in which the random variables are encountered is different to the
previous example. We can readily write down the new $\mathrm{Ne}$,
and the resulting order is given by the trivial permutation $\sigma_{n}[k]=k$.
The function $\mathrm{Pa}$ differs because the permutation does,
but it establishes the same parent relationships as before. This is
not surprising: an SSM can be represented as a directed graphical
model, so that the joint distribution $p_{\sigma_{n}}\left(\mathrm{d}y_{1:T},\mathrm{d}x_{1:T},\mathrm{d}\theta\right)$
associated with each execution $n$ is always the same joint distribution
$p\left(\mathrm{d}y_{1:T},\mathrm{d}x_{1:T},\mathrm{d}\theta\right)$
that appears in (\ref{eq:SSMjoint}), as explained in (\ref{eq:directed-graphical-model}).

This is a simple example. One can imagine more complex code that includes
conditionals (e.g. \texttt{if} statements and \texttt{while} loops)
that may cause only a subset of the random variables to be encountered
on any single execution. The random variables may even be encountered
in different orders, or may be countably infinite (rather than finite)
in number. SSMs that exhibit this complexity occur in, for example,
multiple object tracking. We provide such an example in Section~\ref{sec:worked-example}.

\section{Inference methods for programmatic models\label{sec:methods}}

We wish to infer the conditional distribution of one set of random
variables, given values assigned to some other set. In a Bayesian
context, this amounts to inferring the posterior distribution. For
this purpose, we partition $\{V_{k}\}_{k=1}^{\infty}$ into two disjoint
sets: the \emph{observed} set $O\subseteq\mathcal{P}(\mathbb{N})$
(where $\mathcal{P}$ denotes the power set) containing the indices
of all those random variables for which a value has been given, and
the \emph{latent} set $L\subseteq\mathcal{P}(\mathbb{N})$ with all
other indices, so that $L=\mathbb{N}\setminus O$. We then clamp the
observed random variables to have the given values, i.e. $V_{O}=v_{O}$,
where we use subscript $O$ to select a subset rather than a single
variable. Inference involves computing the posterior distribution:
\begin{align}
\underbrace{p(\mathrm{d}v_{L}\mid v_{O})}_{\text{posterior}} & =\frac{\overbrace{p(v_{O}\mid v_{L})}^{\text{likelihood}}\overbrace{p(\mathrm{d}v_{L})}^{\text{prior}}}{\underbrace{p(v_{O})}_{\text{evidence}}},\label{eq:posterior}
\end{align}
which decomposes into likelihood, prior and evidence terms as annotated.
Having obtained the posterior distribution, we may be interested in
estimating the posterior expectation of some test function of interest,
say $h(V_{L})$:
\[
\mathbb{E}_{p}[h(V_{L})\mid v_{O}]=\int_{\mathbb{V}_{L}}h(v_{L})p(\mathrm{d}v_{L}\mid v_{O}),
\]
and subsequently making decisions based on this result.

A particular execution $n$ of the model code may encounter some subset
of the variables in $O$ and $L$, which we denote $O_{n}$ and $L_{n}$.
The distribution that results from the execution is then:
\begin{align}
p_{\sigma_{n}}(\mathrm{d}v_{L_{n}}\mid v_{O_{n}})\propto & \prod_{k\in L_{n}}p_{\sigma_{n}[k]}\left(\mathrm{d}v_{\sigma_{n}[k]}\mid\mathrm{Pa}(v_{\sigma_{n}[1]},\ldots,v_{\sigma_{n}[k-1]})\right)\times\label{eq:prior}\\
 & \prod_{k\in O_{n}}p_{\sigma_{n}[k]}\left(v_{\sigma_{n}[k]}\mid\mathrm{Pa}(v_{\sigma_{n}[1]},\ldots,v_{\sigma_{n}[k-1]})\right).\label{eq:likelihood}
\end{align}
As before, different executions $m$ and $n$ may yield different
distributions, which may be interpreted as different components of
a mixture. In this case, this mixture is the posterior distribution.

A baseline method for inference is importance sampling from the prior.
The model code is executed: when encountering a random variable in
$L$ it is simulated from the prior, and when encountering a random
variable in $O$ a cumulative weight is updated by multiplying in
the likelihood of the given value. We have then simulated from the
first product (\ref{eq:prior}), and assigned a weight according to
the second product (\ref{eq:likelihood}).

Use of the prior distribution as a proposal is likely to produce estimates
with high variance. We can improve upon this in a number of ways,
such as by maintaining multiple executions simultaneously and selecting
from amongst them in a resampling step, producing a particle filter~\cite{Gordon1993}.
The only limitation here is the alignment of resampling points between
multiple executions in order that resampling is actually beneficial---each
execution may encounter observations in different orders. But because
importance sampling and the most basic particle filters---up to alignment---require
only forward simulation of the prior and pointwise evaluation of the
likelihood, they are unaffected by many of the complexities of programmatic
models, and so particularly suitable for inference. For this reason
they have become a common choice for inference in probabilistic programming
(see e.g. \cite{Wood2014,Paige2014a,Mansinghka2014}). Various optimizations
are available, such as attaching alternative proposal distributions
$q_{\sigma_{n}[k]}$ to random variables in $L$, or marginalizing
out one or more of these. The manual use of these optimizations is
well understood (see \cite{Doucet2011} for a review), although a
key ingredient of probabilistic programming is to automate their use
(see e.g. \cite{Perov2015,Murray2018}). For a tutorial introduction
to the use of the particle filter for nonlinear system identification
we refer to~\cite{SchonLDWNSD:2015,SchonSML:2018}.

Because the structure of a programmatic model may change between executions,
MCMC methods can be difficult to apply. Markov kernels on programmatic
models are, in general, transdimensional, so that techniques such
as reversible jump~\cite{Green1995} are necessary. There are approaches
to automating the design of reversible jump kernels that can work
well in practice (see e.g. \cite{Wingate2011}). Random-walk Metropolis–Hastings
kernels and more-recent gradient-based kernels do not support transdimensional
moves, but might still be applied to the full conditional distribution
of a set of random variables within, for example, a Gibbs sampler.

Particular structural motifs may suggest particular inference methods.
For example, the chain motif suggests the use of specialized Bayesian
filtering methods, while tree and grid motifs are conducive to divide-and-conquer
methods~\cite{Lindsten2017}. Within probabilistic programming, recent
attempts have been made to match inference methods to model substructures,
using both manual and automated techniques (see e.g. \cite{Mansinghka2014,Ge2018,Pfeffer2018,Mansinghka2018}).

The precise choice of inference method depends not only on structure,
but also on form. For example, while the chain motif suggests the
use of a Bayesian filtering method based on structure, the precise
choice of filter depends on form: the Kalman filter for linear-Gaussian
forms, the forward-backward algorithm on HMMs for discrete forms,
the particle filter otherwise. In all cases the structure is the same,
but the form differs. Recognizing this within program code requires
compiler or library support.

Preferably, the choice of inference method based on structure and
form is automated, and ideally by the programming language compiler~\cite{Lunden2018},
which has full access to the abstract syntax tree of the model code
to inspect structure and form. There are fundamental limits to what
can be known at compile time, however. In the general case, at least
some structure and form is unknown until the program is run. For example,
it is not possible to bound the trip count of a loop at compile time~\cite{Nori2014}
if this is a stochastic quantity with unbounded support. The optimal
inference method for a problem may also depend on posterior properties,
such as correlations between random variables, that---by definition---are
unknown \textsl{a priori}. In such situations it may be necessary
to require manual hints provided by the programmer, or to use dynamic
mechanisms that adapt the inference method during execution.

The \emph{delayed sampling}~\cite{Murray2018} heuristic is an example
of the latter. Delayed sampling works for programmatic models in a
similar way to the sum-product algorithm~\cite{Pearl1988} for graphical
models. By keeping a graph of relationships between random variables
as they are encountered by a program, and delaying the simulation
of latent variables for as long as possible, it opens opportunities
for analytical optimizations based on conjugate prior and other relationships.
This includes analytical conditioning, variable elimination, Rao–Blackwellization,
and locally-optimal proposals. It is a heuristic in the sense that
it must make myopic decisions based on the current state of the running
program, without knowledge of its future execution, so that it may
miss potential optimizations. Delayed sampling works through the control
flow statements of a Turing-complete programming language, such as
conditionals and loops, but does not attempt to marginalize over multiple
branches in a single execution.

\subsection{Example}

For the SSM, the task is to infer the latent variables $X_{1:T}$
and $\Theta$ given observations $Y_{1:T}=y_{1:T}$. In a Bayesian
context, this is to infer the posterior distribution
\begin{align}
p(\mathrm{d}x_{1:T},\mathrm{d}\theta\mid y_{1:T})=p(\mathrm{d}x_{1:T}\mid\theta,y_{1:T})p(\mathrm{d}\theta\mid y_{1:T}).\label{eq:SSMposterior}
\end{align}
The first factor in~\eqref{eq:SSMposterior} provides information
about the states. For $t=1,\ldots,T$, its marginals $p(\mathrm{d}x_{t}\mid\theta,y_{1:T})$
are called the filtering distributions, and are the target of Bayesian
filtering methods. The second factor is the target of parameter estimation
methods. We may be interested in obtaining the posterior distribution
over parameters, or obtaining the maximum likelihood estimate instead
by solving the optimization problem
\begin{align}
\widehat{\theta}=\mathrm{argmax}_{\theta}\,p(y_{1:T}\mid\theta).\label{eq:SSMml}
\end{align}
In either case the central object of interest is the likelihood $p(y_{1:T}\mid\theta)$.
By repeated use of conditional probabilities this can be written
\begin{align}
p(y_{1:T}\mid\theta)=\prod_{t=1}^{T}p(y_{t}\mid y_{1:t-1},\theta),\label{eq:SSMlikelihood1}
\end{align}
with the convention that $y_{1:0}=\emptyset$. The terms in the likelihood
are recursively computed via marginalization as
\begin{align}
p(y_{t}\mid y_{1:t-1},\theta) & =\int p(y_{t}\mid x_{t},\theta)p(x_{t}\mid y_{1:t-1},\theta)\,\mathrm{d}x_{t},
\end{align}
so that we obtain
\begin{align}
p(y_{1:T}\mid\theta)=\prod_{t=1}^{T}\int p(y_{t}\mid x_{t},\theta)p(x_{t}\mid y_{1:t-1},\theta)\,\mathrm{d}x_{t}.\label{eq:SSMlikelihood2}
\end{align}

There are numerous ways to compute this likelihood, but for this particular
structure, the family of Bayesian filtering methods is ideal. Within
this family, the preferred method depends on the form of the model.
Recall Code~\ref{code:model}. In general, the functions $f$ and
$g$ must be considered nonlinear, and for such cases the particle
filter is the Bayesian filtering method of choice. Now consider the
code that results from the particular choice $g(x_{t})=x_{t}$: \begin{code}
\begin{lstlisting}[mathescape]
$\theta$ $\sim$ Uniform(0.0, 1.0);
x[1] $\sim$ Gaussian(0.0, 1.0);
y[1] $\sim$ Gaussian(x[1], 0.1);
for t:Integer in 2..10 {
  x[t] $\sim$ Gaussian(f(x[t-1], $\theta$), 1.0);
  y[t] $\sim$ Gaussian(x[t], 0.1);
}
\end{lstlisting}
\caption{\label{code:from-the-particular-choice}}
\end{code}

\noindent A particle filter can still be used, but it is possible
to improve its performance with Rao–Blackwellization and a locally-optimal
proposal (see e.g. \cite{Doucet2011}), by using the local conjugacy
between the Gaussian transition and Gaussian observation models. If
we make the further choice that $f(x_{t-1},\theta)=\theta x_{t-1}$
we have: \begin{code}
\begin{lstlisting}[mathescape]
$\theta$ $\sim$ Uniform(0.0, 1.0);
x[1] $\sim$ Gaussian(0.0, 1.0);
y[1] $\sim$ Gaussian(x[1], 0.1);
for t:Integer in 2..10 {
  x[t] $\sim$ Gaussian($\theta$*x[t-1], 1.0);
  y[t] $\sim$ Gaussian(x[t], 0.1);
}
\end{lstlisting}
\caption{\label{code:linear-gaussian}}
 \end{code}

\noindent On inspection, it is clear that this is now a linear-Gaussian
SSM. In this case we would like to choose the Kalman filter, which
is the optimal Bayesian filtering method for such a form.

Note the important distinction here: the \emph{structure} of the model
is the same in all cases---that of the SSM---and suggests the use of a
Bayesian filtering method over other means to compute the likelihood.
But the \emph{form} of the model differs, and it is these various
forms that suggests the specific Bayesian filtering method to use:
the particle filter, the particle filter with various optimizations,
or the Kalman filter.

\section{Implementation in Birch\label{sec:demonstration}}

In Birch, models are ideally implemented by specifying the joint probability
distribution. Where possible, this means that the code for the model
does not distinguish between latent and observed random variables.
Instead, the value of any random variable can be clamped at runtime
to establish, from the joint distribution, particular conditional
distributions of interest. Methods are also implemented in the Birch
language, rather than being external to the system.

We are particularly interested in considering the structure and form
of a model when choosing a method for inference. This requires some
interface that allows the method implementation to query the model
implementation, and perhaps manipulate its execution. This becomes
complicated with the random structure of programmatic models.

Birch uses a twofold approach. Firstly, meta-programming techniques
are used to represent fine-grain structure and form within the various
mathematical expressions that make up a model. Secondly, the model
programmer has a range of classes available from which they can implement
their model to explicitly describe coarse-grain structure. We deal
with each of these in turn.

\subsection{Fine-grain structure and form}

Birch adopts the meta-programming technique of \emph{expression templates}
to represent mathematical expressions involving random variables.
When such expressions are encountered, they are not evaluated immediately,
but rather arranged into a data structure for later evaluation. In
the meantime, it is possible to inspect and modify that data structure
to facilitate optimizations based on lazy or reordered evaluation.
This provides some of the power of a compiler to inspect structure
and form via an abstract syntax tree, but within the language itself.

Expression templates are common in linear algebra libraries such as
Boost uBLAS and Eigen~\cite{Guennebaud2010}, where they are used
to omit unnecessary memory allocations, reads and writes, and so produce
more efficient code. They are also used in reverse-mode automatic
differentiation to compute the gradient of a function, such as in
Stan for HMC, and to implement delayed sampling. At this stage, they
are primarily used in Birch for this latter purpose.

Recall Code~\ref{code:model}; declarations were omitted but might
be as follows: \begin{code}
\begin{lstlisting}[mathescape]
$\theta$:Real;
x:Real[T];
y:Real[T];
\end{lstlisting}
\caption{\label{code:might-be-as-follows}}
\end{code}This declares \texttt{$\theta$}, and the arrays \texttt{x} and
\texttt{y}, as being ordinary variables of type \texttt{Real}. We
can instead declare them as random variates of type \texttt{Real}
like so: \begin{code}
\begin{lstlisting}[mathescape]
$\theta$:Random<Real>;
x:Random<Real>[T];
y:Random<Real>[T];
\end{lstlisting}
\caption{\label{code:like-so}}
\end{code}

\noindent Various mathematical functions and operators are overloaded
for this type \texttt{Random\textless{}Real\textgreater{}}, to construct
a data structure for lazy evaluation, rather than eager evaluation.

\begin{figure}[p]
\includegraphics[width=1\textwidth]{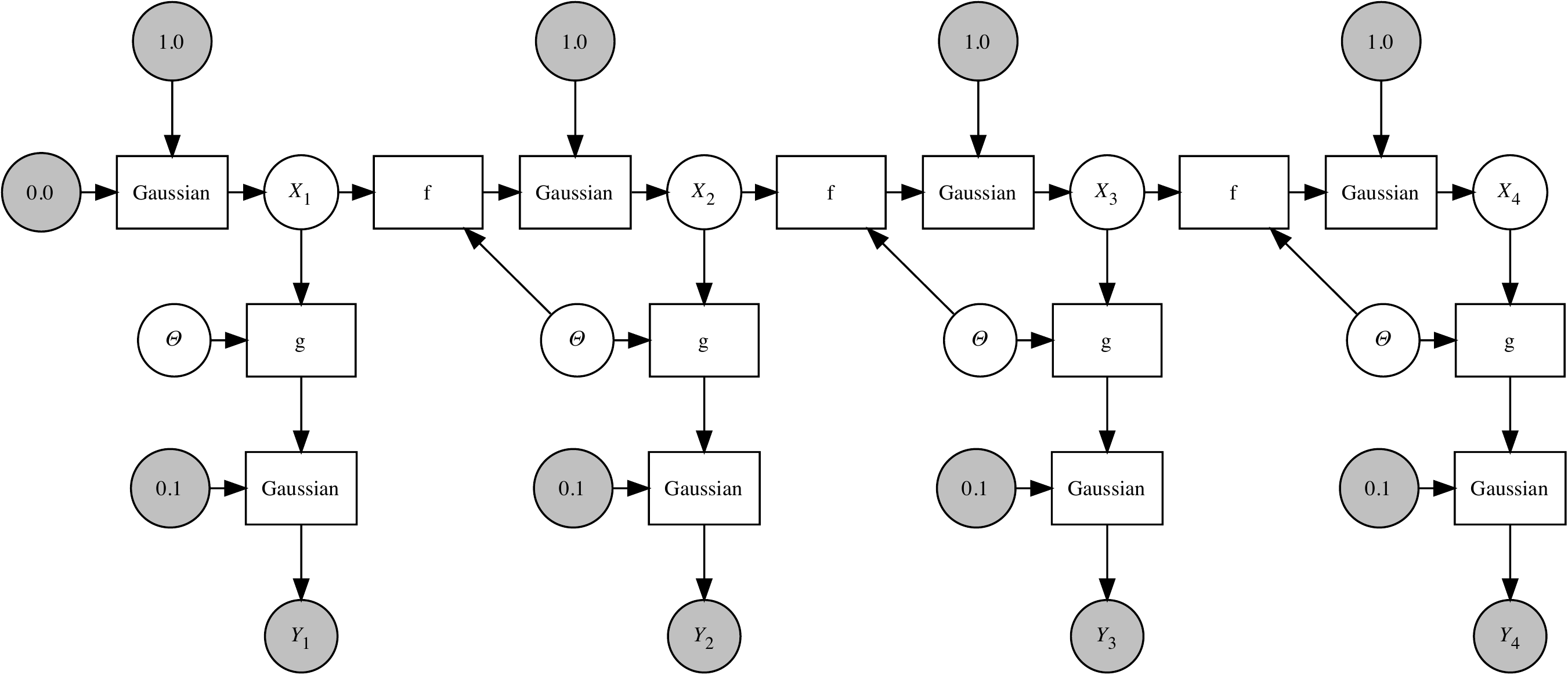} \caption{A data structure describing Code~\ref{code:model} after four iterations.
Shaded circles represent literals and observed random variables. Empty
circles represent latent random variables. Rectangles indicate function
calls, with inbound edges denoting their arguments and outbound edges
their result. The node $\Theta$ has been repeated for clarity.\label{fig:ExpressionTemplate}}
\end{figure}

\begin{figure}[p]
\includegraphics[width=1\textwidth]{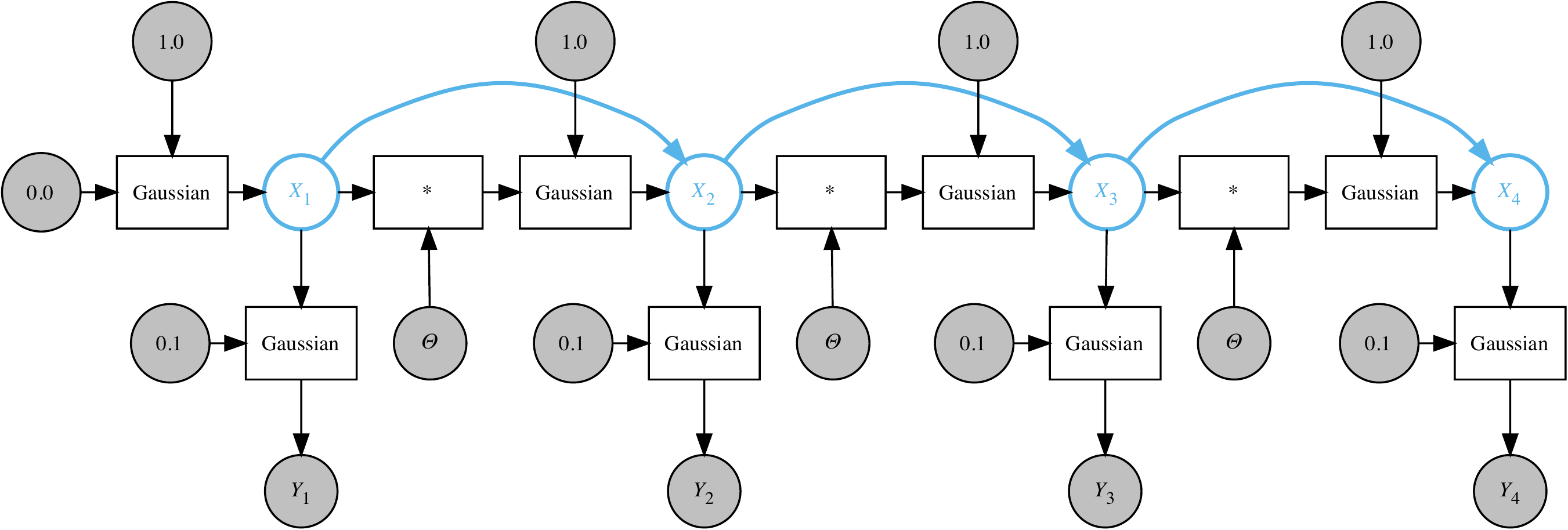} \caption{As Figure \ref{fig:ExpressionTemplate} but for Code~\ref{code:linear-gaussian}.
Here, the delayed sampling mechanism in Birch recognizes the linear
relationships between a chain of Gaussian random variables (highlighted),
and can apply appropriate optimizations based on analytical marginalization
and conditioning.\label{fig:ExpressionTemplate2}}
\end{figure}

When Code~\ref{code:model} is executed using these random types,
a data structure such as that in Figure~\ref{fig:ExpressionTemplate}
is constructed. This represents all the functions that are called,
and their arguments, without evaluating them until necessary. As the
functions $f$ and $g$ are opaque, there is little of value to discover
here at first. If, however, we were to code the model with $f(x_{t-1},\theta)=\theta x_{t}$
and $g(x_{t})=x_{t}$, as in Code~\ref{code:linear-gaussian}, we
would obtain the data structure in Figure~\ref{fig:ExpressionTemplate2}.
Now, Birch identifies a chain of Gaussian random variables for which
it is able to marginalize and condition forward analytically, using
closed form solutions with which it has been programmed. This path
is annotated in Figure~\ref{fig:ExpressionTemplate2} (it is precisely
the $M$-path used in delayed sampling~\cite{Murray2018}). Ultimately,
the computations performed are identical to running a Kalman filter
forward, followed by recursively sampling backward. This utilizes
structure and form to yield an exact sample from the posterior distribution,
using a method that is more efficient that other means.

\subsection{Coarse-grain structure}

Birch provides abstract classes for various structural motifs and
model classes. These automate some of the task of assembling a model
by encapsulating boilerplate code, and provide crucial information
on coarse-grain structure that can be used by an inference method.
Consider, again, Code~\ref{code:model}. A more complete implementation,
as a class, may look something like this: \begin{code}
\begin{lstlisting}[mathescape]
class Example < Model {
  a:Random<Real>;
  x:Random<Real>[10];
  y:Random<Real>[10];

  fiber simulate() -> Real {
    a $\sim$ Uniform(0.0, 1.0);
    x[1] $\sim$ Gaussian(0.0, 1.0);
    y[1] $\sim$ Gaussian(x[1], 0.1);
    for t:Integer in 2..10 {
      x[t] $\sim$ Gaussian(a*x[t-1], 1.0);
      y[t] $\sim$ Gaussian(x[t], 0.1);
    }
  }
}
\end{lstlisting}
\caption{\label{code:more-complete-implementation}}
\end{code}

\noindent This declares a new class called \texttt{Example} that inherits
from the class \texttt{Model}, provided by the Birch standard library.
At this stage, classes in Birch use a limited subset of the functionality
of the classes in C++ to which they are compiled. They support single
but not multiple inheritance, and all member functions are virtual.
Code \ref{code:more-complete-implementation} makes use of a \emph{fiber}
(also called a \emph{coroutine}). Intuitively, this is a function
for which execution can be paused and resumed. Each time execution
is paused, the fiber \emph{yields} a value to the caller, in a manner
analogous to the way that a function \emph{returns} a value to its
caller---although a fiber may yield many times while a function returns
only once. Like member functions, member fibers are virtual in Birch.

The \texttt{Example} class in Code \ref{code:more-complete-implementation}
declares some member variables to represent the random variables of
the model, then specifies the joint probability distribution over
them by overriding the \texttt{simulate} fiber inherited from \texttt{Model}.
This fiber simply simulates the model forward, but does so incrementally
via implicit yields in the $~$ statements,
which have a particular behavior. If the random variable on the left
has not yet been assigned a value, the $~$
statement associates it with the distribution on the right, so that
a value can be simulated for it later. If, on the other hand, the
random variable on the left has previously been assigned a value,
the $~$ statement computes its log-likelihood
under the distribution on the right and yields the result. This yield
value is always of type \texttt{Real}, as shown in the fiber declaration.
This forms the most basic interface by which an inference method can
incrementally evaluate likelihoods and posteriors, and even condition
by assigning values to random variables in the model before executing
the \texttt{simulate} fiber. It also represents the ideal of writing
code to specify the joint distribution, while assigning values to
variables before execution to simulate from a conditional distribution
instead.

It is clear that the model represented by Code \ref{code:more-complete-implementation}
has the structure of an SSM as in (\ref{eq:SSMjoint}). A compiler
with sophisticated static analysis may recognize this. Lacking such
sophistication, it is the responsibility of the programmer to provide
some hints. The choice to inherit from the \texttt{Model} class provides
no such hints---the model is essentially a black box. Alternatives do
provide hints. The Birch standard library provides a generic class
called\texttt{ StateSpaceModel} that itself inherits from \texttt{Model},
but reveals more information about structure to an inference method,
and handles the boilerplate code for such structure. \texttt{StateSpaceModel}
is a generic class that takes a number of additional type arguments
to specify the type of the parameters, state variables, and observed
variables. A class that inherits from \texttt{StateSpaceModel} should
also override the \texttt{parameter}, \texttt{initial}, \texttt{transition},
and \texttt{observation} member fibers to describe the components
of the model. These four fibers replace the \texttt{simulate} member
fiber that is overridden when inheriting directly from \texttt{Model}.

The implementation may look something like Code \ref{more-complete-implementation-structured}.
Type arguments are given to \texttt{StateSpaceModel\textless{}...\textgreater{}}
in the inheritance clause to indicate that the parameters, state variables
and observed variables are all of type \texttt{Random\textless{}Real\textgreater{}}.
The \texttt{x'} that appears in the \texttt{transition} fiber is simply
a name; the prime is a valid character for names, motivated by bringing
the representation in code closer to the representation in mathematics.

\begin{code}[htp]
\begin{lstlisting}[mathescape]
class Example < StateSpaceModel<Random<Real>,Random<Real>,Random<Real>> {
  fiber parameter(a:Random<Real>) -> Real {
    a $\sim$ Uniform(0.0, 1.0);
  }
  fiber initial(x:Random<Real>, a:Random<Real>) -> Real {
    x $\sim$ Gaussian(0.0, 1.0);
  }
  fiber transition(x':Random<Real>, x:Random<Real>, a:Random<Real>) -> Real {
    x' $\sim$ Gaussian(a*x, 1.0);
  }
  fiber observation(y:Random<Real>, x:Random<Real>, a:Random<Real>) -> Real {
    y $\sim$ Gaussian(x, 0.1);
  }
}
\end{lstlisting}
\caption{\label{more-complete-implementation-structured}}
\end{code}

The model structure defined through the \texttt{StateSpaceModel} class
is a straightforward extension of the sorts of interfaces that existing
software provides. LibBi~\cite{Murray2015}, for example, is based
entirely on the SSM structure, and the user implements their model
by writing code blocks with the same four names as the fibers above.
But while LibBi supports only one interface, for SSMs, Birch can support
many interfaces, for many model classes, with an object-oriented approach.

\section{Example in Birch\label{sec:worked-example}}

We demonstrate the above ideas on a multiple object tracking problem
(see e.g. \cite{Vo2015} for an overview). Such problems arise in
application domains including air traffic control, space debris monitoring
(e.g. \cite{Jones2015}), and cell tracking (e.g. \cite{Ulman2017}).
In all of these cases, some unknown number of objects appear, move,
and disappear in some physical space, with noisy sensor measurements
of their positions during this time. The task is to recover the object
tracks from these noisy sensor measurements.

The model to be introduced is an SSM within the class of programmatic
models described in Section \ref{sec:models}. The size of the latent
state $X_{t}$ varies according to the number of objects, which is
unknown \textsl{a priori}, and furthermore changes in time as objects
appear and disappear. Similarly, the size of the observation $Y_{t}$
varies according to the number of objects and, in addition, rates
of detection and noise. While it is straightforward to simulate from
the joint distribution $p(\mathrm{d}x_{1:T},\mathrm{d}y_{1:T})$,
simulation of the posterior distribution $p(\mathrm{d}x_{1:T}\mid y_{1:T})$
is complicated by a \emph{data association} problem: the random matching
of observations to objects, which includes both missing and spurious
detections (clutter). That is, the structure of the relationships
between the components of $X_{t}$ and $Y_{t}$ is not fixed. For
$M$ observations and $K$ detected objects, there are $(M+1)^{(K)}$
(rising factorial) possible associations of equal probability under
the prior. Naive inference with forward simulation of this association,
as in a bootstrap particle filter, will not scale beyond a small number
of objects and observations. Optimizations that instead leverage the
structure and form of the model are necessary.

For this example, we work on a two-dimensional rectangular domain
with lower corner $l=(-10,-10)$ and upper corner $u=(10,10)$. The
model is described---and ultimately implemented---in a hierarchical way,
by first specifying a model for single objects, then combining several
of these into a model for multiple objects.

\subsection{Single object model}

The single object model describes the dynamics and observations (including
missing detections) of a single object. The state of that object is
represented by a six-dimensional vector giving its position, velocity
and acceleration in the two-dimensional space. These evolve according
to a linear-Gaussian SSM. Using superscripts to denote object $i$,
the initial and transition models are given by:
\begin{align*}
X_{0}^{i} & \sim\mathcal{N}(\mu_{0}^{i},M), & X_{t}^{i}\sim\mathcal{N}(Ax_{t-1}^{i},Q),
\end{align*}
where $\mu_{0}^{i}$ has its position component drawn uniformly on
the domain $[l,u]$, with its velocity and acceleration components
set to zero, and $M$, $A$ and $Q$ are the matrices:
\begin{align*}
M & =\left(\begin{array}{ccc}
5I & 0 & 0\\
0 & 0.1I & 0\\
0 & 0 & 0.01I
\end{array}\right) & A & =\left(\begin{array}{ccc}
I & I & 0.5I\\
0 & I & I\\
0 & 0 & I
\end{array}\right) & Q & =\left(\begin{array}{ccc}
0 & 0 & 0\\
0 & 0 & 0\\
0 & 0 & 0.01I
\end{array}\right),
\end{align*}
with $I$ the $2\times2$ identity matrix and $0$ the $2\times2$
zero matrix.

At time $t$, the object may or may not be detected. This is indicated
by the variable $D_{t}^{i}$, which takes value 1 with probability
$\rho$ to indicate detection, and value 0 otherwise. If detected,
the observation is of position only:
\begin{align*}
D_{t}^{i} & \sim\mathrm{Bernoulli}(\text{\ensuremath{\rho}}), & Y_{t}^{i} & \sim\mathcal{N}(Bx_{t}^{i},R),
\end{align*}
with matrices:

\begin{align*}
B & =\left(\begin{array}{ccc}
I & 0 & 0\end{array}\right), & R & =0.1I.
\end{align*}

\subsection{Multiple object model}

The multiple object model mediates several single object models, including
their appearance and disappearance, and clutter. At time $t$, the
number of new objects to appear, $B_{t}$, is distributed as:
\[
B_{t}\sim\mathrm{Poisson}(\lambda)
\]
for some parameter $\lambda$. The lifetime of each object, $S^{i}$,
is distributed as:
\[
S^{i}\sim\mathrm{Poisson}(\tau)
\]
for some parameter $\tau$. In practice this is implemented as a probability
of disappearance at each time step. If object $i$ has been present
for $s^{i}$ time steps, its probability of disappearing on the next
is given by $\Pr[S^{i}=s^{i}]/\Pr[S^{i}\geq s^{i}]$, with these probabilities
easily computed under the Poisson distribution.

At time $t$, some number of spurious observations (clutter) occur
that are not associated with an object. Their number is denoted $C_{t}$,
distributed as
\begin{equation}
C_{t}-1\sim\mathrm{Poisson}(\mu)\label{eq:clutter}
\end{equation}
for some parameter $\mu$. The position of each is uniformly distributed
on the domain $[l,u]$. That there is at least one spurious observation
at each time merely simplifies the implementation slightly---we revisit
this point in Section \ref{sec:implementation} below.

\subsection{Inference method\label{sec:inference}}

Inference can leverage the structure and form of the model. The structure
is that of an SSM, with random latent state size and random observation
size. Within this SSM is further structure, as each of the single
objects follows an SSM independently of the others. The form of those
inner SSMs is linear-Gaussian, suggesting the use of a Kalman filter
for optimal tracking, while the outer SSM requires the use of a particle
filter. The inference problem is further complicated by data association,
specifically matching detected objects with given observations. This
is handled in the multiple object model, with a specific proposal
distribution used within the particle filter to propose associations
of high likelihood.

Let $O_{t}^{i}$ denote the index of the observation associated with
object $i$ at time $t$. For an object $i$ that is not detected
($D_{t}^{i}=0$) we set $O_{t}^{i}=0$. If there are $N_{t}$ number
of objects, of which $K_{t}=\sum_{i=1}^{N_{t}}d_{t}^{i}$ are detected,
and $M_{t}$ number of observations, the prior distribution over associations
is uniform on the $(M_{t}+1)^{(K_{t})}$ possibilities:
\[
p(\mathrm{d}o_{t}^{1:N_{t}}\mid d_{t}^{1:N_{t}})=\prod_{i=1}^{N_{t}}p(\mathrm{d}o_{t}^{i}\mid o_{t}^{1:i-1},d_{t}^{i}),
\]
where the pmfs associated with the conditional distributions on the
right are:
\[
p(O_{t}^{i}=j\mid o_{t}^{1:i-1},d_{t}^{i})=\begin{cases}
\frac{\mathds{1}[j\notin\{o_{t}^{1},\ldots,o_{t}^{i-1}\}]}{\sum_{m=1}^{M_{t}}\mathds{1}[m\notin\{o_{t}^{1},\ldots,o_{t}^{i-1}\}]} & \text{\text{if }}d_{t}^{i}=1\text{ and }j\in\{1,\ldots,M_{t}\},\\
1 & \text{if }d_{t}^{i}=0\text{ and }j=0,\\
0 & \text{otherwise.}
\end{cases}
\]
Here, $\mathds{1}$ denotes the indicator function. These expressions
simply limit the uniform distribution to the correct domain: as long
as all $O_{t}^{1:N_{t}}$ are in the support and the nonzero $O_{t}^{1:N_{t}}$
(corresponding to detected objects) are distinct, the probability
is uniformly $1/(M_{t}+1)^{(K_{t})}$.

The proposal distribution is to iterate through the detected objects
in turn, choosing for each an associated observation in proportion
to its likelihood, excluding those observations already associated.
Thus, we have:
\[
q(\mathrm{d}o_{t}^{1:N_{t}}\mid d_{t}^{1:N_{t}})=\prod_{i=1}^{N_{t}}q(\mathrm{d}o_{t}^{i}\mid o_{t}^{1:i-1},d_{t}^{i}),
\]
where the pmfs associated with the conditional distributions on the
right are:
\[
q(O_{t}^{i}=j\mid o_{t}^{1:i-1},d_{t}^{i})=\begin{cases}
\frac{p(y_{t}^{j}\mid x_{t}^{i})\mathds{1}[j\notin\{o_{t}^{1},\ldots,o_{t}^{i-1}\}]}{\sum_{m=1}^{M_{t}}p(y_{t}^{m}\mid x_{t}^{i})\mathds{1}[m\notin\{o_{t}^{1},\ldots,o_{t}^{i-1}\}]} & \text{\text{if }\ensuremath{d_{t}^{i}}=1}.\\
1 & \text{if }d_{t}^{i}=0\text{ and }j=0\\
0 & \text{otherwise.}
\end{cases}
\]
The delayed sampling heuristic within Birch automatically applies
a further optimization to this. As each object follows a linear-Gaussian
SSM, the $X_{t}^{i}$ are marginalized out using a Kalman filter,
so that the proposal becomes:
\[
q^{*}(O_{t}^{i}=j\mid o_{t}^{1:i-1},d_{t}^{i})=\begin{cases}
\frac{\phi(y_{t}^{j};\,\hat{y}_{t}^{i},\hat{\Sigma}_{t}^{i})\mathds{1}[j\notin\{o_{t}^{1},\ldots,o_{t}^{i-1}\}]}{\sum_{m=1}^{M_{t}}\phi(y_{t}^{m};\,\hat{y}_{t}^{i},\hat{\Sigma}_{t}^{i})\mathds{1}[m\notin\{o_{t}^{1},\ldots,o_{t}^{i-1}\}]} & \text{\text{if }\ensuremath{d_{t}^{i}}=1}.\\
1 & \text{if }d_{t}^{i}=0\text{ and }j=0\\
0 & \text{otherwise,}
\end{cases}
\]
where $\phi$ is the pdf of the multivariate normal distribution,
with $\hat{y}_{t}^{i}$ the mean and $\hat{\Sigma}_{t}^{i}$ the covariance
of $Y_{t}^{i}$ as predicted by the Kalman filter for the $i$th object.

At time $t$, then, it is straightforward to propose a data association
$o_{t}^{1:N_{t}}$ from $q$ (or $q^{*}$) and, in the usual importance
sampling fashion, weight with the ratio:
\begin{align*}
\frac{p(o_{t}^{1:N_{t}}\mid d_{t}^{1:N_{t}})}{q(o_{t}^{1:N_{t}}\mid d_{t}^{1:N_{t}})} & =\frac{(M_{t}+1)^{(K_{t})}}{\prod_{i=1}^{N_{t}}q(o_{t}^{i}\mid o_{t}^{1:i-1},d_{t}^{i})}.
\end{align*}
Unassociated observations at the end of the procedure are considered
clutter.

\subsection{Implementation\label{sec:implementation}}

The model is implemented in Birch in a modular fashion. It consists
of three classes: one for the parameters, one for the single object
model, and one for the multiple object model.

A class called \texttt{Global} is declared with a member variable
for each of the parameters of the model, given in Code \ref{code:global}.

\begin{code}[tph]
\begin{lstlisting}[mathescape]
class Global {
  l:Real[_];    // lower corner of domain of interest
  u:Real[_];    // upper corner of domain of interest
  d:Real;       // probability of detection
  M:Real[_,_];  // initial value covariance
  A:Real[_,_];  // transition matrix
  Q:Real[_,_];  // state noise covariance
  B:Real[_,_];  // observation matrix
  R:Real[_,_];  // observation noise covariance
  $\lambda$:Real;        // birth rate
  $\mu$:Real;        // clutter rate
  $\tau$:Real;        // track length rate
}
\end{lstlisting}
\caption{Parameters of the multiple object tracking model in Birch.\label{code:global}}
\end{code}

The implementation of the single object model is given in Code \ref{code:single-object-model}.
This declares a class called \texttt{Track} that, as before, inherits
from the class \texttt{StateSpaceModel} in the Birch standard library.
The parameter type is \texttt{Global}, while the state and observation
types are both \texttt{Random\textless{}Real{[}\_{]}\textgreater{}}.
The \texttt{initial}, \texttt{transition}, and \texttt{observation}
member fibers are overridden to specify the model. An unfamiliar operator
appears in the code: the meaning of $<\sim$ is to simulate from the distribution on the right and assign the value
to the variable on the left.

\begin{code}[t]
\begin{lstlisting}[mathescape]
class Track < StateSpaceModel<Global,Random<Real[_]>,Random<Real[_]>> {
  t:Integer;  // starting time of the track

  fiber initial(x:Random<Real[_]>, $\theta$:Global) -> Real {
    auto $\mu$ <- vector(0.0, 3*length($\theta$.l));
    $\mu$[1..2] $<\sim$ Uniform($\theta$.l, $\theta$.u);
    x $\sim$ Gaussian($\mu$, $\theta$.M);
  }

  fiber transition(x':Random<Real[_]>, x:Random<Real[_]>, $\theta$:Global) -> Real {
    x' $\sim$ Gaussian($\theta$.A*x, $\theta$.Q);
  }

  fiber observation(y:Random<Real[_]>, x:Random<Real[_]>, $\theta$:Global) -> Real {
    d:Boolean;
    d $<\sim$ Bernoulli($\theta$.d);  // is the track detected?
    if d {
      y $\sim$ Gaussian($\theta$.B*x, $\theta$.R);
    }
  }
}
\end{lstlisting}
\caption{Implementation of the single object model in Birch.\label{code:single-object-model}}
\end{code}

The implementation of the multiple object model is given in Code \ref{code:multiple-object-model}.
This declares a class called \texttt{Multi} that, again, inherits
from the class \texttt{StateSpaceModel} in the Birch standard library.
The parameter type is given as \texttt{Global}, the state type as
a \texttt{List} of \texttt{Track} objects, and the observation type
as a \texttt{List} of \texttt{Random\textless{}Real{[}\_{]}\textgreater{}}
objects; \texttt{List} is a generic class provided by the standard
library. The \texttt{transition} and \texttt{observation} member fibers
are overridden to specify the model. The \texttt{observation} fiber
is complicated by the data association problem. Recall, as in (\ref{eq:clutter}),
that there is always at least one point of clutter; an empty list
of observations can therefore be interpreted as missing observations
to be simulated, rather than present observations to be conditioned
upon. When observations are present, the code defers to an alternative
\texttt{association} member fiber, detailed below. When missing, they
are simulated. This is an instance where it is not possible to write
a single piece of code that specifies the joint distribution and covers
both cases.

\begin{code}[tp]
\begin{lstlisting}[mathescape]
class Multi < StateSpaceModel<Global,List<Track>,List<Random<Real[_]>>> {
  t:Integer <- 0;  // current time

  fiber transition(x':List<Track>, x:List<Track>, $\theta$:Global) -> Real {
    t <- t + 1;

    /* move current objects */
    auto track <- x.walk();
    while track? {
      $\rho$:Real <- pmf_poisson(t - track!.t - 1, $\theta$.$\tau$);
      R:Real <- 1.0 - cdf_poisson(t - track!.t - 1, $\theta$.$\tau$) + $\rho$;
      s:Boolean;
      s $<\sim$ Bernoulli(1.0 - $\rho$/R);  // does the object survive?
      if s {
        track!.step();
        x'.pushBack(track!);
      }
    }

    /* birth new objects */
    N:Integer;
    N $<\sim$ Poisson($\theta$.$\lambda$);
    for n:Integer in 1..N {
      track:Track;
      track.t <- t;
      track.$\theta$ <- $\theta$;
      track.start();
      x'.pushBack(track);
    }
  }

  fiber observation(y:List<Random<Real[_]>>, x:List<Track>, $\theta$:Global) -> Real {
    if !y.empty() {  // observations given, use data association
      association(y, x, $\theta$);
    } else {
      N:Integer;
      N $<\sim$ Poisson($\theta$.$\mu$);
      for n:Integer in 1..(N + 1) {
        clutter:Random<Real[_]>;
        clutter $<\sim$ Uniform($\theta$.l, $\theta$.u);
        y.pushBack(clutter);
      }
    }
  }
}
\end{lstlisting}
\caption{Implementation of the multiple object model in Birch.\label{code:multiple-object-model}}
\end{code}

Finally, we show the \texttt{association} member fiber, which is the
most difficult part of the model and inference method. This is given
in Code \ref{code:association}. A number of new language features
appear. First, the \texttt{yield} statement is used to yield a value
from a fiber, much like the \texttt{return} statement is used to return
a value from a function. In previous examples, yielding from fibers
has been implicit, via $~$ operators, rather
than explicit, via \texttt{yield} statements. Here, explicit yields
are used to correct weights for the data association proposal described
in Section \ref{sec:inference}. Another unfamiliar operator appears:
the meaning of $\sim>$ is to compute
the pdf or pmf of the value of the variable on the left under the
distribution on the right, and implicitly yield its logarithm. The
symmetry with the previously-introduced $<\sim$
operator is intentional: these two operators work as a pair for simulation
and observation. Other than this, the code makes use of the interface
to the \texttt{Random\textless{}Real{[}\_{]}\textgreater{}} class
to query for detection and compute likelihoods.

\begin{code}[tp]
\begin{lstlisting}[mathescape]
  fiber association(y:List<Random<Real[_]>>, x:List<Track>, $\theta$:Global) -> Real {
    K:Integer <- 0;  // number of detections
    auto track <- x.walk();
    while track? {
      if track!.y.back().hasDistribution() {
        /* object is detected, compute proposal */
        K <- K + 1;
        q:Real[y.size()];
        n:Integer <- 1;
        auto detection <- y.walk();
        while detection? {
          q[n] <- track!.y.back().pdf(detection!);
          n <- n + 1;
        }
        Q:Real <- sum(q);

        /* propose an association */
        if Q > 0.0 {
          q <- q/Q;
          n $<\sim$ Categorical(q);  // choose an observation
          yield track!.y.back().realize(y.get(n));  // likelihood
          yield -log(q[n]);  // proposal correction
          y.erase(n);  // remove the observation for future associations
        } else {
          yield -inf;  // detected, but all likelihoods (numerically) zero
        }
      }

      /* factor in prior probability of hypothesis */
      yield -lrising(y.size() + 1, K);  // prior correction
    }

    /* clutter */
    y.size() - 1 $\sim>$ Poisson($\theta$.$\mu$);
    auto clutter <- y.walk();
    while clutter? {
      clutter! $\sim>$ Uniform($\theta$.l, $\theta$.u);
    }
  }
\end{lstlisting}
\caption{Implementation of the data association step in Birch.\label{code:association}}
\end{code}

\subsection{Results}

\begin{figure}[t]
\begin{minipage}[t]{0.32\columnwidth}%
\includegraphics[width=1\textwidth]{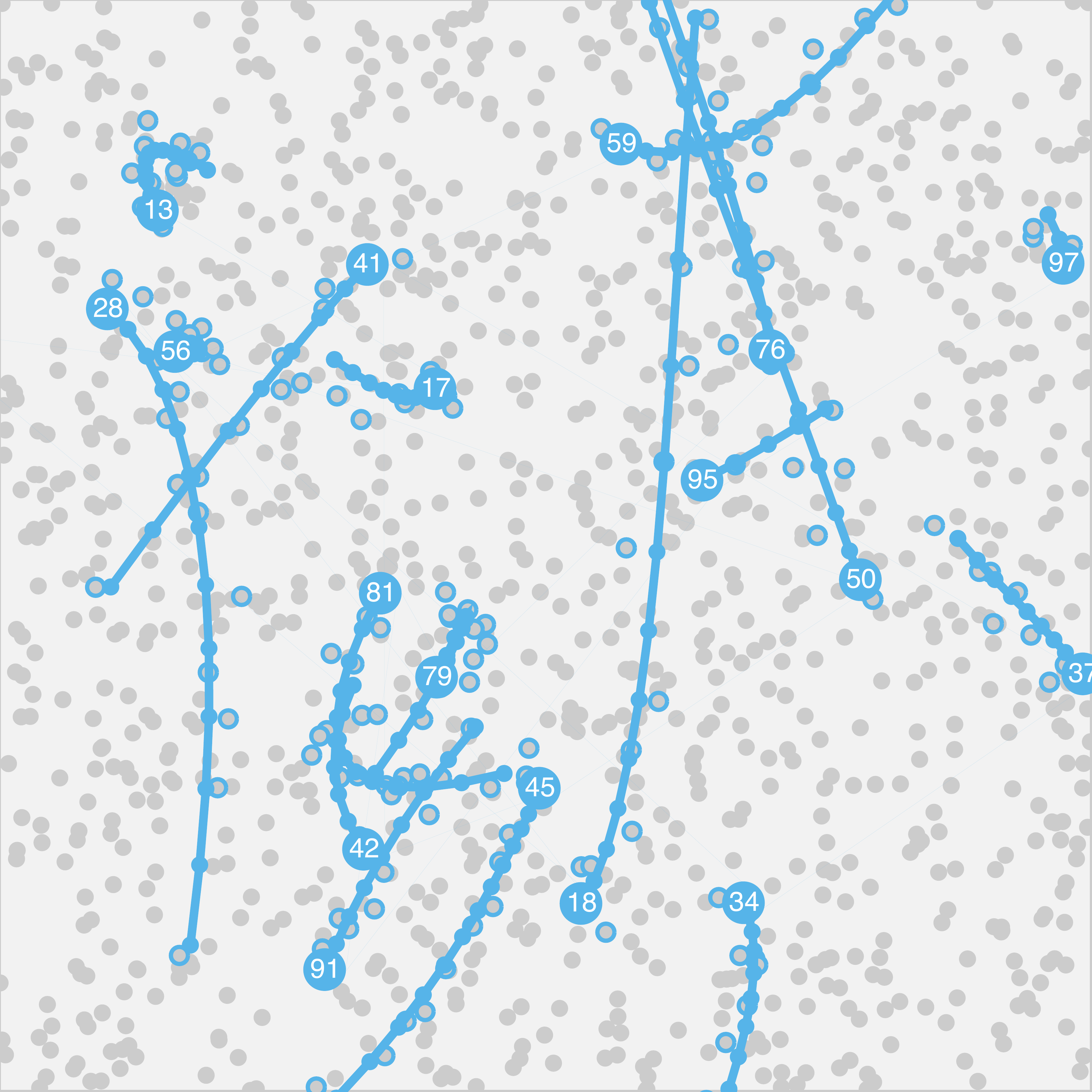}%
\end{minipage}\hfill{}%
\begin{minipage}[t]{0.32\columnwidth}%
\includegraphics[width=1\textwidth]{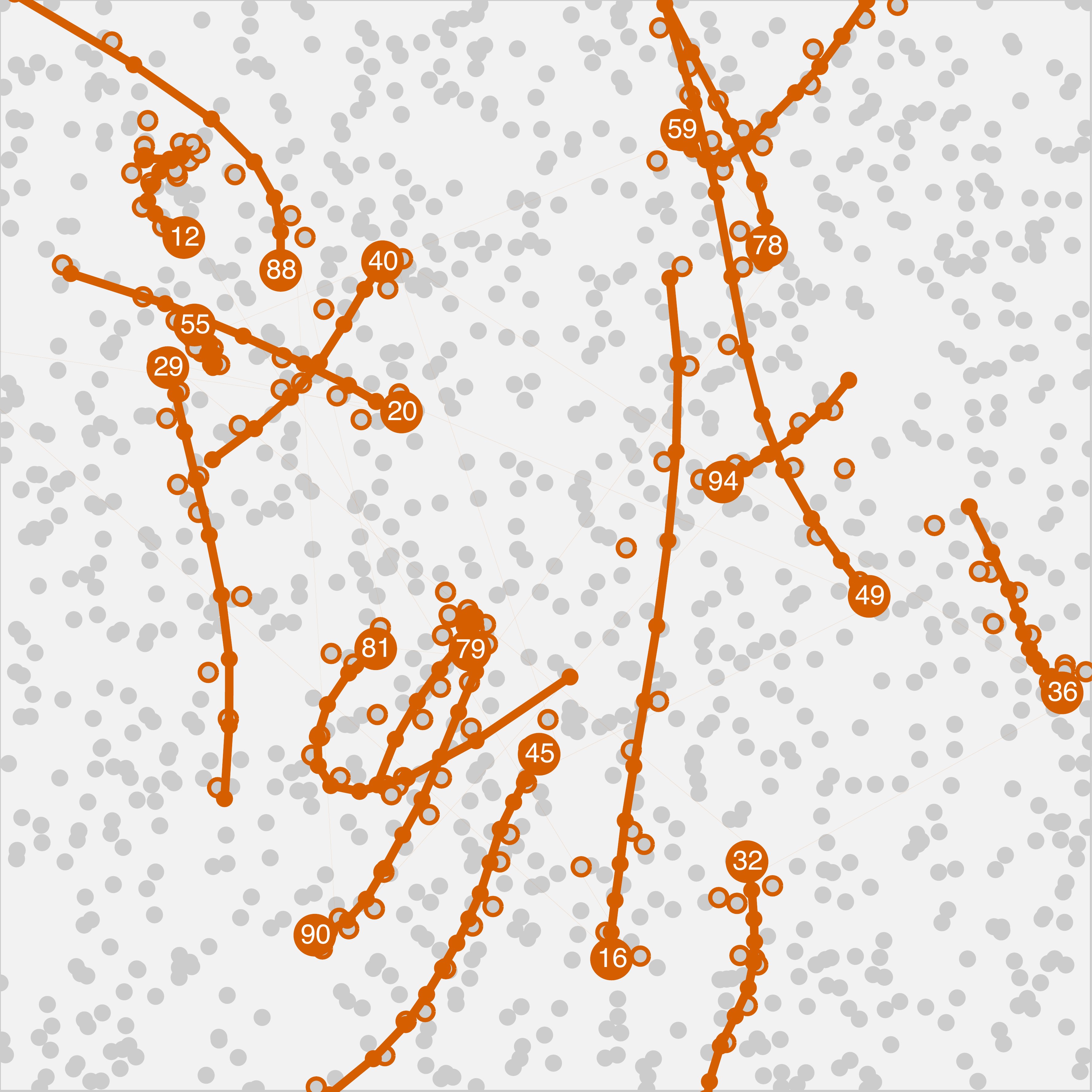}%
\end{minipage}\hfill{}%
\begin{minipage}[t]{0.32\columnwidth}%
\includegraphics[width=1\textwidth]{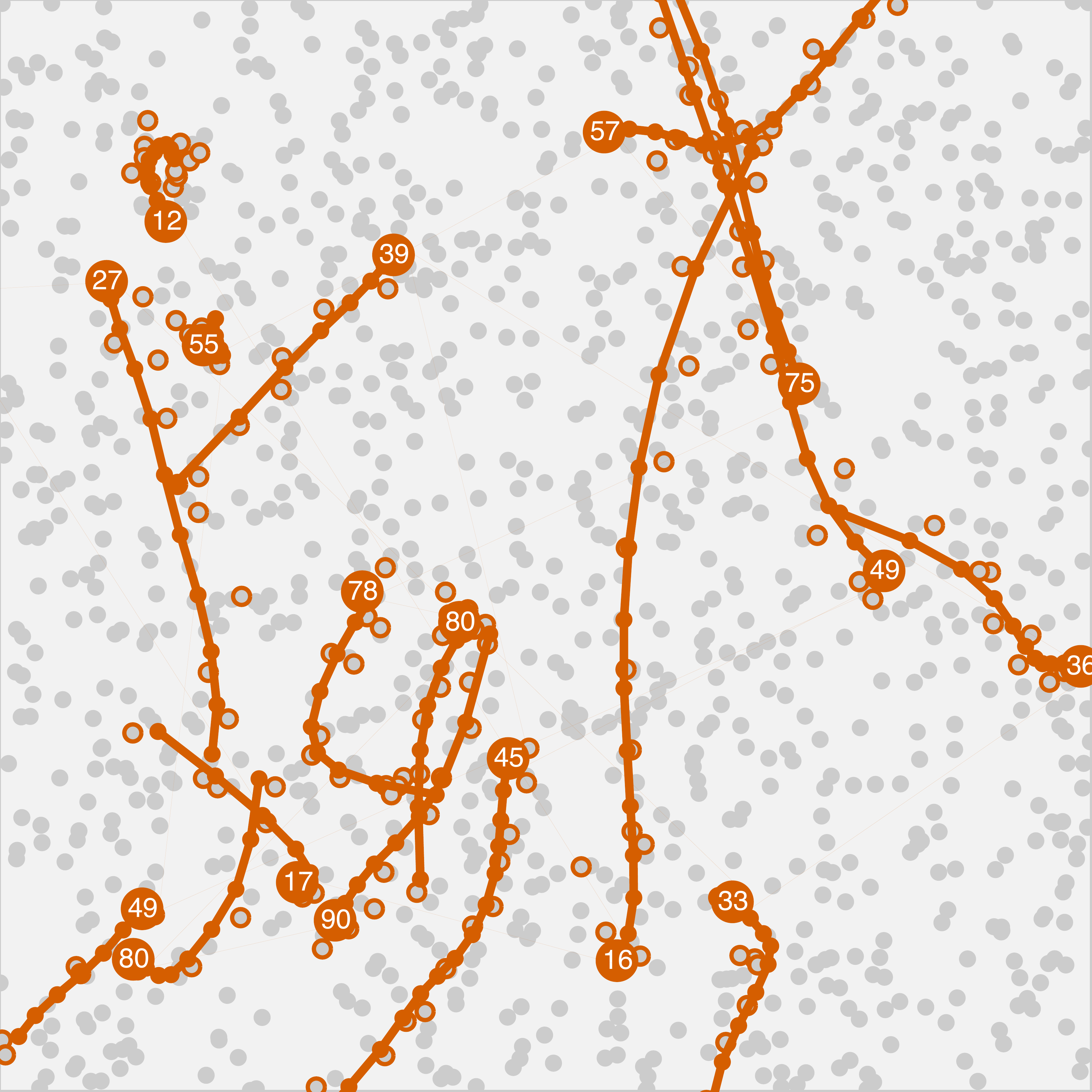}%
\end{minipage}\vspace{2mm}

\begin{minipage}[t]{0.32\columnwidth}%
\includegraphics[width=1\textwidth]{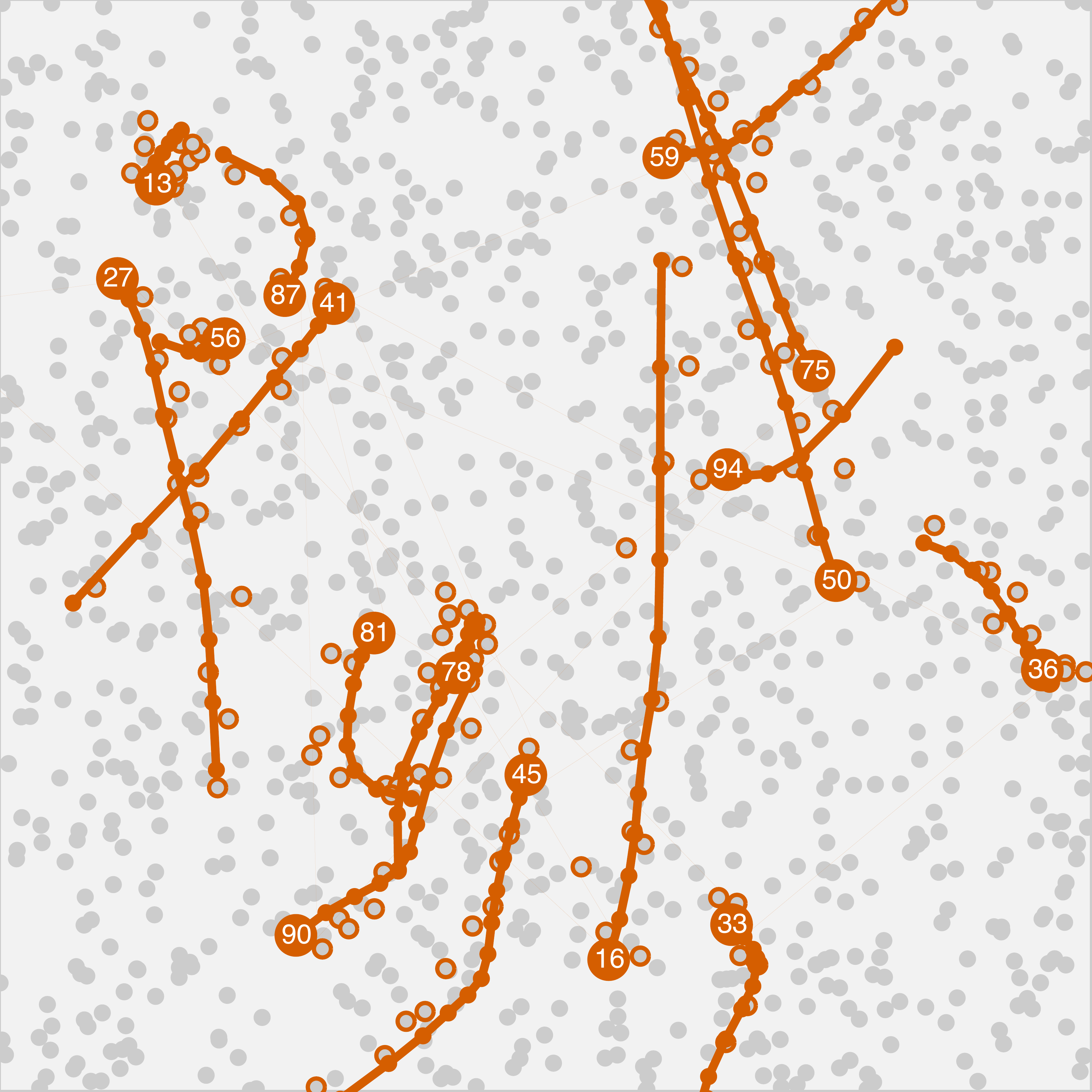}%
\end{minipage}\hfill{}%
\begin{minipage}[t]{0.32\columnwidth}%
\includegraphics[width=1\textwidth]{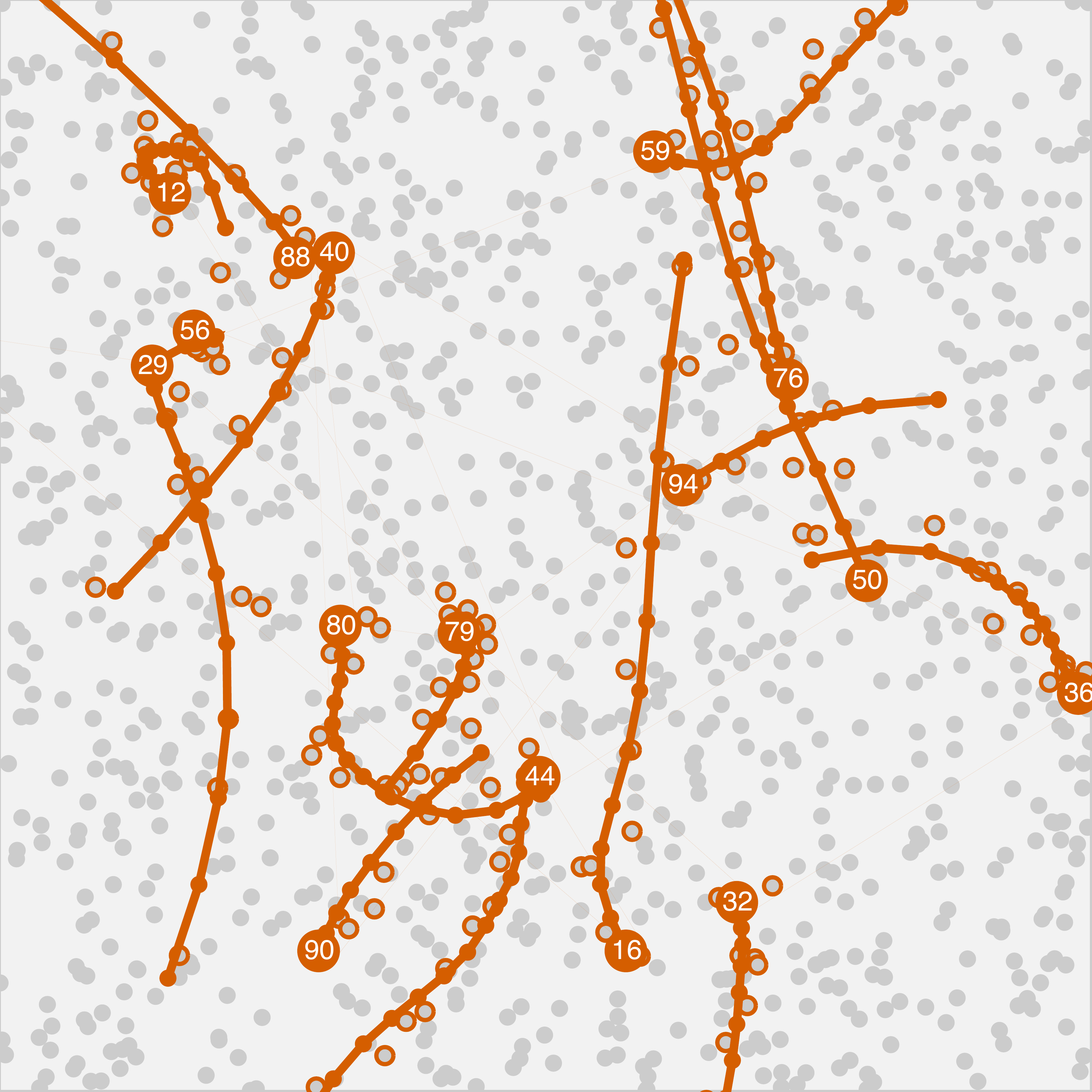}%
\end{minipage}\hfill{}%
\begin{minipage}[t]{0.32\columnwidth}%
\includegraphics[width=1\textwidth]{filter5}%
\end{minipage}

\caption{Phase plots for the multiple object tracking example, showing ground
truth (top left) and posterior samples (others) with associated log-weights
(left to right, top to bottom) of -7868.9, -7864.6, -7873.6, -7865.4,
-7870.3. Lines represent object tracks. Points along the lines represent
the position at each time. The larger point at the start of each line
denotes the birth position, labelled with the birth time. Grey dots
represent observations; those circled are associated with an object
track, while others are classified as clutter.\label{fig:multiple-object-tracking}}

\end{figure}

The model is simulated for 100 time steps to produce the ground truth
and data set shown in the top left of Figure \ref{fig:multiple-object-tracking}.
Using Birch, we then run a particle filter several times to produce
the remaining plots in Figure \ref{fig:multiple-object-tracking}.
The delayed sampling feature of Birch ensures that, within each particle
of the particle filter, a separate Kalman filter is applied to each
object. Each run uses 32768 particles, from which a single path is
drawn as a posterior sample, to be weighted by the normalizing constant
estimate obtained from the particle filter.

Generally, these posterior samples show good alignment with the ground
truth. The longer tracks in the posterior samples correspond to similar
tracks in the ground truth. Some shorter tracks in the ground truth
are missing in the posterior samples (for example, the short track
in the top right of the ground truth that appears at time 97), and
conversely, some spurious tracks that do not appear in the ground
truth appear sporadically in some posterior samples. These spurious
tracks should be expected: the prior puts positive probability on
objects being detected very few times---or not at all---in their lifetime.

\section{Summary\label{sec:discussion}}

Probabilistic programming is a relatively young field that seeks to
accelerate the workflow of probabilistic modeling and inference using
new probabilistic programming languages. A key concept is to decouple
the implementation of models and inference methods, using various
techniques to match them together for efficient inference, preferably
in an automated way.

In this work, we have provided a definition of the class of \emph{programmatic
models}, with an emphasis on \emph{structure} and \emph{form}. The
class reflects the nature of models expressed in programming languages
where, in general, structure and form are not known \textsl{a priori},
but may instead depend on random choices made during program execution.
We have discussed some of the complexity that this brings to inference,
especially that different executions of the model code may encounter
different sets of random variables. SMC methods have issues around
alignment for resampling, where different particles may encounter
observations in different orders. MCMC methods encounter issues around
Markov kernels needing to be transdimensional. Nevertheless, we argue
that persistent substructure is common, can be represented by structural
motifs, and can be utilized in implementation to match models with
appropriate inference methods.

We have shown the particular implementation of these ideas in the
universal probabilistic programming language Birch. Here, expression
templates are used to explore fine-grain structure and form, while
class interfaces reveal coarse-grain structure. The expression templates
in particular are important to enable analytical optimizations via
the delayed sampling heuristic.

Finally, we have shown a multiple object tracking example, where the
model involved resides in the class of programmatic models, featuring
a random number of latent variables, random number of observed variables,
and random associations between them. Nonetheless, the model exhibits
a clear structure and form: multiple linear-Gaussian SSMs for single
objects, within a single nonlinear SSM for multiple objects. The delayed
sampling heuristic provided by Birch automatically adapts the inference
method to a particle filter for the multiple object model, with each
particle within that filter using a Kalman filter for each single
object model.

\section{Acknowledgements}

This research was financially supported by the Swedish Foundation
for Strategic Research (SSF) via the project \emph{ASSEMBLE} (contract
number: RIT15-0012). The authors thank Karl Granström for helpful
discussions around the multiple object tracking example.

\bibliographystyle{abbrvnat}
\bibliography{birch}

\end{document}